\documentclass[10pt,twocolumn,letterpaper]{article}

\usepackage[pagenumbers]{cvpr}

\usepackage{adjustbox}
\usepackage{colortbl}
\usepackage{tabularx}
\usepackage{multirow}
\usepackage{float} 
\definecolor{revisionbg}{RGB}{240,240,240}

\makeatletter
\def\fps@figure{t}
\def\fps@table{t}
\makeatother

\AtBeginDocument{%
    \crefname{equation}{Eq.}{Eqs.}%
    \Crefname{equation}{Eq.}{Eqs.}%
    \crefname{table}{Tab.}{Tabs.}%
    \Crefname{table}{Tab.}{Tabs.}%
    \crefname{figure}{Fig.}{Figs.}%
    \Crefname{figure}{Fig.}{Figs.}%
}

\definecolor{cvprblue}{rgb}{0.21,0.49,0.74}
\usepackage[pagebackref,breaklinks,colorlinks,allcolors=cvprblue]{hyperref}

\def\confName{CVPR}
\def\confYear{2027}

\title{When to Trust Adaptation: Contextual Fusion for\\
Transferable Vision-Language Prompting}

\newcommand{\cafauthorblock}{%
\begin{minipage}{\dimexpr\textwidth-2\tabcolsep\relax}
\centering
{\normalsize Zhenyuan Chen$^{1}$ \quad Wenshuo Tao$^{1}$ \quad Zhuo Huang$^{4}$ \quad
Shuo Chen$^{5}$ \quad Ming-Ming Cheng$^{2,1}$ \quad Xiang Li$^{2,1,3}$\thanks{Corresponding author: Xiang Li.}\par}
\vspace{3pt}
{\fontsize{9}{10.5}\selectfont
$^{1}$VCIP, College of Computer Science, Nankai University, China;
$^{2}$NKIARI, Shenzhen Futian, Nankai University, China;\\
$^{3}$AAIS, Nankai University, China;
$^{4}$Australian Institute for Machine Learning;\\
$^{5}$School of Intelligence Science and Technology, Nanjing University, China.\par}
\vspace{3pt}
{\small\ttfamily zhenyuanchen@mail.nankai.edu.cn; xiang.li.implus@nankai.edu.cn\par}
\end{minipage}%
}
\author{\cafauthorblock}

\makeatletter
\patchcmd{\@maketitle}{\vspace*{24pt}}{\vspace*{14pt}}{}{\PackageError{caf}{Title spacing patch failed}{}}
\patchcmd{\@maketitle}{\vspace*{12pt}}{\vspace*{6pt}}{}{\PackageError{caf}{Author spacing patch failed}{}}
\makeatother

\hypersetup{
    pdftitle={When to Trust Adaptation: Contextual Fusion for Transferable Vision-Language Prompting},
    pdfauthor={Zhenyuan Chen, Wenshuo Tao, Zhuo Huang, Shuo Chen, Ming-Ming Cheng, Xiang Li},
    pdfkeywords={Vision-Language Models, Prompt Learning, Adaptive Fusion, Cross-Dataset Transfer, Base-to-Novel Generalization}
}

\begin{document}
\maketitle
\begingroup
\setlength{\textfloatsep}{8pt plus 1pt minus 1pt}
\setlength{\dbltextfloatsep}{8pt plus 1pt minus 1pt}
\setlength{\floatsep}{8pt plus 1pt minus 1pt}
\setlength{\dblfloatsep}{8pt plus 1pt minus 1pt}
\setlength{\intextsep}{8pt plus 1pt minus 1pt}
\captionsetup{skip=4pt}
\begin{abstract}
When a vision--language model encounters a new class or domain, how much should it trust its learned prompts?
Because source adaptation can correct pretrained errors on some target images while disturbing valid predictions on others, we introduce Contextual Adaptive Fusion (CAF), which retains four prediction branches by pairing frozen and prompted visual and textual representations.
To arbitrate when branches disagree on an image, CAF combines each branch's predictive confidence with how frequently the branches favor its proposed class across the unlabeled target collection, placing greater trust in predictions that are locally confident yet less dominated by collection-wide frequency bias.
The fusion coefficients are learned once on source data and remain fixed at inference.
Trained on ImageNet, CAF achieves 69.96\% average accuracy across ten cross-dataset benchmarks (1.92 percentage points above ManiPT) and remains highly competitive across four ImageNet variants, setting a new best of 52.21\% on ImageNet-S.
In base-to-novel transfer across eleven datasets, CAF further reaches 79.77\% novel-class accuracy and an 82.81\% harmonic mean.
Code is available at \url{https://github.com/zhenyuanchenai/CAF}.
\end{abstract}
    
\section{Introduction}
\label{sec:intro}

\suppressfloats[t]
\begin{figure}[t]
    \centering
    \includegraphics[width=\columnwidth]{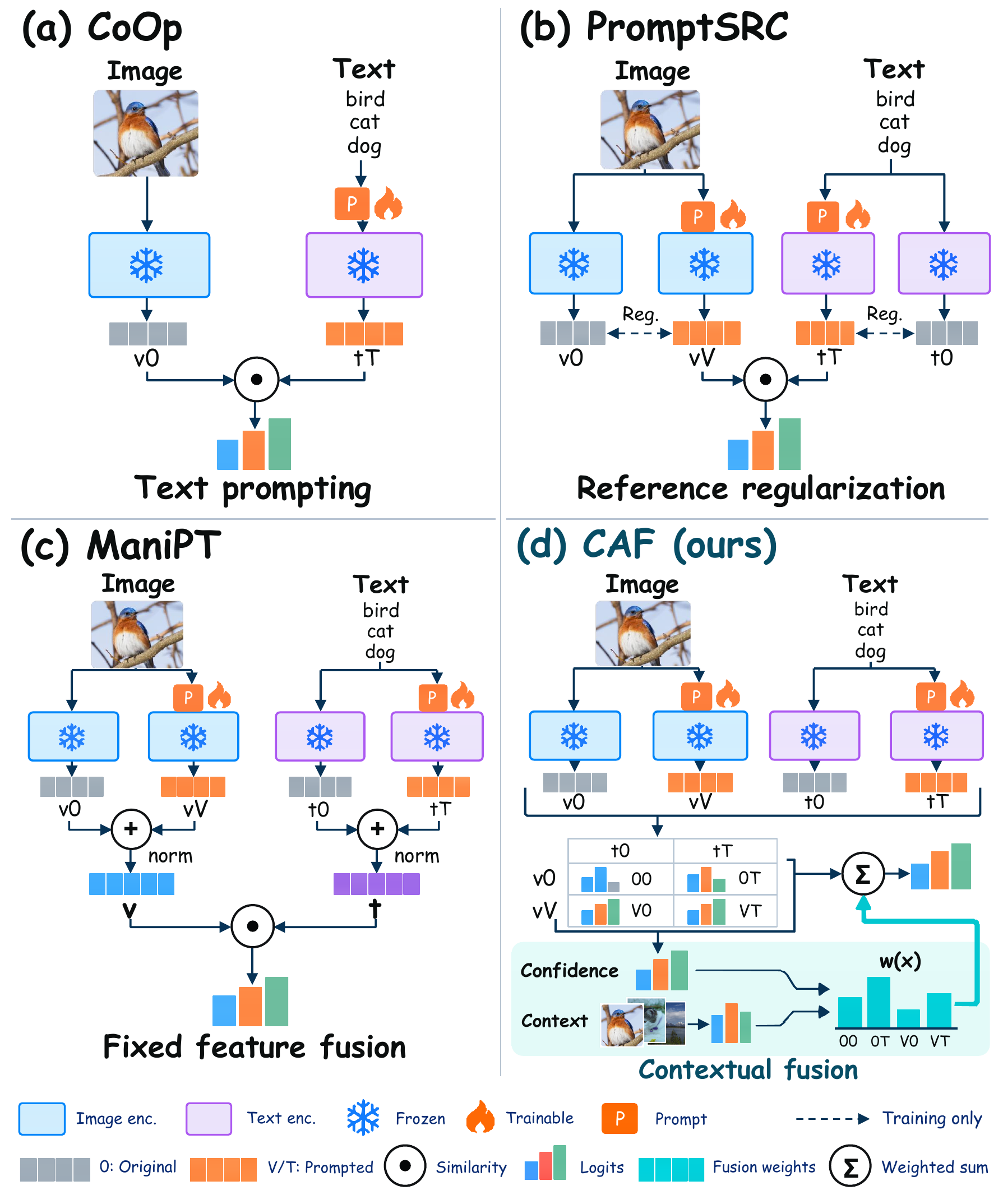}
    \caption{Comparison of CLIP prompting strategies: (a) text prompting in CoOp~\cite{zhou2022learning}; (b) frozen-reference regularization in PromptSRC~\cite{khattak2023self}; (c) fixed feature fusion in ManiPT~\cite{yang2026prompt}; and (d) contextual fusion in CAF, where instance-level confidence and collection-level context determine image-dependent weights for four prediction branches. All backbone encoders remain frozen.}
    \label{fig:method_comparison}
\end{figure}

Vision-language models such as CLIP~\cite{radford2021learning} enable visual recognition through text descriptions of candidate categories.
As illustrated by CoOp in Fig.~\ref{fig:method_comparison}(a), prompt learning adapts these models with a small set of trainable tokens while keeping the pretrained encoders frozen~\cite{zhou2022learning,khattak2023maple}.
Transferable vision-language prompting aims to adapt pretrained models with limited labeled data while preserving their ability to generalize to new images, categories, and visual domains~\cite{ren2023prompt,chen2024revisiting}.
However, source adaptation can improve recognition of seen classes while degrading performance on unseen classes, even relative to the frozen pretrained model~\cite{zhou2022conditional}.
A central challenge is therefore to exploit source supervision without sacrificing the pretrained model's generalizable recognition ability.

To address this challenge, existing methods regularize adaptation against frozen references~\cite{khattak2023self,roy2024consistency,guo2025mmrl,guo2026mmrl,ma2026alignednorm} (Fig.~\ref{fig:method_comparison}(b)).
ManiPT~\cite{yang2026prompt} further retains pretrained information at inference through fixed fusion of frozen and prompted features (Fig.~\ref{fig:method_comparison}(c)).
Because multimodal prompt learning adapts both visual features and textual class representations~\cite{khattak2023maple,khattak2023self,zheng2025hierarchical,rahman2025dimple,c2025osloprompt}, retaining their frozen counterparts provides four possible image--text pairings and hence four prediction branches.
Fusing these four branches is challenging because the relative reliability of pretrained and adapted predictions can vary across images~\cite{zhu2024robust}, while shifts in visual distributions or class vocabularies can further complicate the transfer of source-derived fusion rules~\cite{zhou2022conditional,zhu2024robust}.
Consequently, fusion weights optimized for source-domain accuracy can compromise robustness under distribution shift, as the optimal balance between pretrained and adapted predictions may differ across distributions~\cite{zhu2024robust}.
The central question is therefore \emph{when to trust adaptation}: how can we assign appropriate contributions to these four branches for each target image without access to target labels?

Determining these contributions requires evidence of each branch's relative reliability for the current image~\cite{ding2024weak,zhu2024robust,rodriguezopazo2025synergy}.
Predictive confidence characterizes how certain a branch is about the current image, while predictions over the unlabeled target collection reveal which classes the branches tend to favor.
We use this collective prediction behavior as a common reference for interpreting each branch's confidence, linking the assessment of an individual prediction to the statistical context of its proposed class.
This motivates two complementary types of features for adaptive fusion: \emph{instance-level confidence features} and \emph{collection-level contextual features}.
We instantiate these feature types with normalized predictive entropy~\cite{shu2022test} and a shared prediction-frequency statistic, respectively.
The latter is computed by averaging class probabilities across unlabeled target images and all four branches, with each branch reading the statistic associated with its proposed class.
Together, these features provide a contextual basis for deciding when to trust adaptation, allowing branch contributions to depend on both image-specific confidence and the collective prediction tendencies of the target collection.

Building on these two feature types, we propose \textbf{Contextual Adaptive Fusion (CAF)}, which combines frozen and prompted predictions through image-dependent weighting (Fig.~\ref{fig:method_overview}).
CAF first learns independent visual and textual prompts with frozen encoders, using source classification together with regularization for visual margin preservation and textual prototype alignment.
Using the resulting prompted representations alongside their frozen counterparts, CAF evaluates all four image--text pairings to retain both pretrained and adapted predictions for fusion.
During source training, we learn two scalar fusion coefficients from labeled source examples to balance instance-level confidence and collection-level context. The fusion loss updates only these coefficients, while a separate objective trains the prompts.
These coefficients are held fixed across target datasets, while the features they combine are computed from the target predictions.
At inference, CAF follows a transductive protocol, pooling class probabilities over the full unlabeled target collection and all four branches to obtain the shared prediction-frequency statistic.
For each image, this statistic supplies a negative-log-frequency correction at each branch's top-ranked class, which complements its negative normalized entropy when the branches disagree.
The two terms are combined using the learned coefficients, and the resulting scores are normalized with a softmax to produce image-dependent weights for fusing the original branch logits.
This procedure produces predictions over the full target class vocabulary without a trainable scoring network, target labels, or target-time parameter updates.

Trained on ImageNet, CAF achieves state-of-the-art (SOTA) cross-dataset transfer with 69.96\% average accuracy across ten benchmarks, surpassing ManiPT~\cite{yang2026prompt} by 1.92 percentage points and ranking first on eight datasets.
CAF also achieves the best accuracy on ImageNet-S at 52.21\%, exceeding ManiPT by 1.58 points, and remains competitive across the four ImageNet variants.
In base-to-novel transfer across eleven datasets, CAF achieves the highest average novel-class accuracy (79.77\%) and harmonic mean (82.81\%), surpassing ManiPT by 1.10 and 0.72 percentage points, respectively.

Our main contributions are summarized as follows.
\begin{itemize}
    \item \textbf{Four-Branch Transfer Formulation.} We formulate transferable prompting over four pairings of frozen and prompted visual--textual representations, decoupling unimodal and joint adaptation.

    \item \textbf{Contextual Adaptive Fusion (CAF).} We dynamically weight the four branches per image by combining branch confidence with target-collection prediction frequency via two fixed, source-learned coefficients.

    \item \textbf{Mechanistic and Empirical Insights.} We show that CAF concentrates its gains on images with branch disagreement, adapts modality preferences across transfer regimes, and achieves SOTA transfer.
\end{itemize}

\begin{figure*}[!t]
    \centering
    \includegraphics[width=\textwidth]{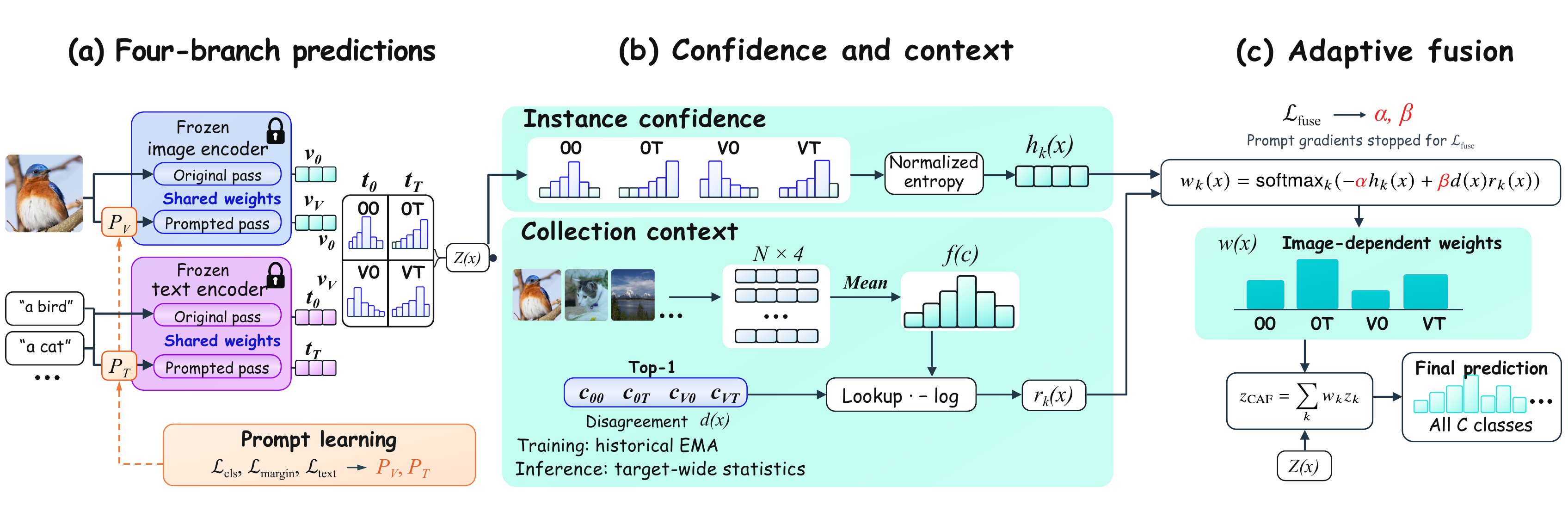}
    \caption{\textbf{Overview of CAF.} (a) Frozen and prompted visual and textual representations form four prediction branches. (b) Normalized entropy measures instance-level uncertainty, while class probabilities averaged across images and branches provide collection-level context. (c) Two source-learned coefficients combine these signals into image-dependent weights that fuse the original logits over all classes. Prompt learning and fusion-coefficient learning use separate gradients; all learned parameters remain fixed at inference.}
    \label{fig:method_overview}
\end{figure*}

\section{Related Work}
\label{sec:related-work}

\subsection{Transferable Vision-Language Prompting}

Prompt learning adapts pretrained vision-language models such as CLIP~\cite{radford2021learning} through a small set of trainable tokens while keeping the backbone encoders frozen.
Early approaches learn textual contexts~\cite{zhou2022learning}, condition prompts on individual images~\cite{zhou2022conditional}, or introduce visual prompts~\cite{jia2022visual}, while MaPLe~\cite{khattak2023maple} jointly adapts the visual and textual modalities via cross-modal prompt coupling.
Subsequent methods enrich prompt learning with class attributes~\cite{tian2024argue,ding2025tree,li2025advancing}, language-model knowledge~\cite{zheng2024large}, task-aware context~\cite{hao2025task}, and hierarchical cross-modal interactions~\cite{zheng2025hierarchical}.
Other approaches improve the learned representations through feature disentanglement~\cite{rahman2025dimple}, category- and sample-level confusion modeling~\cite{shao2026capt}, and spectral guidance~\cite{zhou2026specpl}.
Prompt diversity is explored through variational modeling~\cite{Derakhshani2022VariationalPT}, surrogate feature generation~\cite{pmlr-v267-liu25bx}, and repulsive Bayesian sampling~\cite{bendou2026rebapl}.
Complementary approaches expand the knowledge available to prompt learning through large-vocabulary pretraining~\cite{ren2023prompt}, teacher-guided distillation~\cite{li2024promptkd}, and cascaded prompt boosting and adaptation~\cite{wu2024cascade}.
Beyond source-time prompt learning, TPT~\cite{shu2022test} adapts prompts at test time through entropy minimization with confidence-based selection of augmented views.

Preserving pretrained knowledge during adaptation is another central direction for generalizable prompt learning.
ProGrad~\cite{zhu2023prompt} constrains prompt updates using gradients derived from pretrained predictions, whereas KgCoOp~\cite{yao2023visual} regularizes learned textual embeddings toward those obtained from hand-crafted prompts.
PromptSRC and CoPrompt~\cite{khattak2023self,roy2024consistency} encourage consistency with pretrained representations, and SPTR~\cite{cui2025similarity} combines textual regularization with a similarity-based formulation to retain knowledge from hand-crafted prompts.
More recent methods, including MMRL, MMRL++, and AlignedNorm~\cite{guo2025mmrl,guo2026mmrl,ma2026alignednorm}, combine multimodal adaptation with regularization informed by pretrained representations.
ManiPT~\cite{yang2026prompt} further retains frozen features alongside prompted features at inference.
CAF builds on regularized multimodal prompting to learn adapted representations while retaining their frozen counterparts for prediction fusion.
It complements representation-level knowledge preservation with contextual weighting of four prediction branches, using instance-level confidence and collection-level context.

\subsection{Prediction Fusion for Transferable Adaptation}

Combining pretrained and adapted knowledge provides a complementary route to preserving transferability.
WiSE-FT~\cite{wortsman2022robust} interpolates the parameters of zero-shot and fine-tuned models, while Tip-Adapter~\cite{zhang2022tip} augments zero-shot logits with cache-based predictions retrieved from labeled support examples using a global mixing coefficient.
ManiPT~\cite{yang2026prompt} combines frozen and prompted features through fixed normalized addition before computing image--text similarities.
These approaches incorporate complementary information through global mixing coefficients or fixed feature addition.

Adaptive fusion determines model contributions using input-dependent evidence.
Ding et al.~\cite{ding2024weak} compare the in-distribution scores of zero-shot and prompt-tuned classifiers to obtain sample-dependent fusion weights.
Variance Reduction Fine-tuning (VRF)~\cite{zhu2024robust} uses proximity to a reference set of zero-shot failures to adjust the contribution of the fine-tuned model.
In multi-backbone ensembling, Rodriguez-Opazo et al.~\cite{rodriguezopazo2025synergy} learn input-dependent temperature coefficients from visual representations to combine predictions from different CLIP backbones.
These approaches derive their weighting signals from classifier outputs, reference-set proximity, or visual representations.

The unlabeled target collection provides an additional source of information for adapting predictions.
TransCLIP~\cite{zanella2024boosting} exploits the structure of this collection through transductive inference regularized by pretrained vision-language predictions.
CAF incorporates collection-level information into the weighting of four prediction branches formed from frozen and prompted representations.
It combines instance-level confidence features with collection-level contextual features, using predictive entropy and shared soft prediction frequencies to assess individual predictions in their target context.
The two fusion coefficients are learned during source training and remain fixed at target inference, while the shared prediction frequencies are computed from the full unlabeled target collection.
CAF thus performs transductive prediction fusion through a directly computed weighting rule, without a trainable routing network or target-time parameter updates.

\section{Method}
\label{sec:method}

\subsection{Problem Setup and Four-Branch Predictions}
\label{sec:problem_setup}

We consider transferable vision-language prompting from a labeled source dataset $\mathcal D_s$ with class vocabulary $\mathcal C_s$ to an unlabeled target collection $\mathcal U_t$ with vocabulary $\mathcal C_t$.
The visual distributions and class vocabularies may differ across source and target.
Target class names and unlabeled images are available for transductive inference, without target labels or parameter updates.

As illustrated in Fig.~\ref{fig:method_overview}(a), CAF learns independent visual and textual prompts while keeping both encoders frozen.
We denote the frozen and prompted image features by $v_0(x)$ and $v_V(x)$, and their textual counterparts by $t_0(c)$ and $t_T(c)$.
All four representations are $\ell_2$-normalized.
Following ManiPT and CoPrompt~\cite{yang2026prompt,roy2024consistency}, $t_0(c)$ is formed by averaging normalized embeddings of fixed GPT-generated class descriptions and renormalizing the result.
The prompted feature $t_T(c)$ encodes the class name with learned textual prompts.

CAF retains four image--text pairings, indexed by $a\in\{0,V\}$ and $b\in\{0,T\}$:
\begin{equation}
    z_{ab}(x,c)=\tau\,v_a(x)^\top t_b(c),
    \label{eq:four_branches}
\end{equation}
where $\tau$ is the frozen pretrained logit scale.
The indices $a$ and $b$ specify the visual and textual states: $0$ denotes an unprompted representation, and $V$ or $T$ denotes prompting.
This yields four branches, $\mathcal K=\{00,0T,V0,VT\}$, which share the same class ordering and logit scale and reuse the two visual and two textual representations.
Their full-vocabulary logits provide the prediction sources for subsequent contextual fusion.

\subsection{Regularized Multimodal Prompt Learning}
\label{sec:prompt_learning}

CAF learns visual and textual prompts through source classification while regularizing each modality against its frozen reference.
For a source minibatch $\mathcal B$, classification supervision is applied to the fully prompted $VT$ branch:
\begin{equation}
    \mathcal L_{\mathrm{cls}}
    =\frac{1}{|\mathcal B|}
    \sum_{(x,y)\in\mathcal B}
    \operatorname{CE}\!\left(z_{VT}(x),y\right).
    \label{eq:source_classification}
\end{equation}

To preserve reliable pretrained decisions during visual adaptation, we define the true-class margin as $m_k(x,y)=z_k(x,y)-\max_{c\in\mathcal C_s\setminus\{y\}}z_k(x,c)$.
When the frozen $00$ branch has a positive margin, we penalize any reduction in the corresponding $V0$ margin:
\begin{equation}
    \begin{aligned}
        \mathcal L_{\mathrm{margin}}
        &=\frac{1}{|\mathcal B|}
        \sum_{(x,y)\in\mathcal B}
        \mathbf 1\!\left\{m_{00}(x,y)>0\right\}\\
        &\quad{}\times
        \left[m_{00}(x,y)-m_{V0}(x,y)\right]_+,
    \end{aligned}
    \label{eq:margin_preservation}
\end{equation}
where $[a]_+=\max(a,0)$.
Comparing $00$ and $V0$ keeps the textual reference fixed, thereby constraining the effect of visual prompting.

For textual adaptation, we align the prompted representations with the frozen class prototypes over the entire source vocabulary:
\begin{equation}
    \mathcal L_{\mathrm{text}}
    =\frac{1}{|\mathcal C_s|}
    \sum_{c\in\mathcal C_s}
    \left[1-t_T(c)^\top t_0(c)\right].
    \label{eq:text_alignment}
\end{equation}
Since both representations are normalized, this term penalizes their angular discrepancy and anchors textual adaptation to the pretrained semantic references.

The overall prompt-learning objective is
\begin{equation}
    \mathcal L_{\mathrm{prompt}}
    =\mathcal L_{\mathrm{cls}}
    +\lambda_{\mathrm{margin}}\mathcal L_{\mathrm{margin}}
    +\lambda_{\mathrm{text}}\mathcal L_{\mathrm{text}},
    \label{eq:prompt_objective}
\end{equation}
where the two coefficients control the regularization strengths.
Classification updates both prompt sets, while margin preservation and prototype alignment update only the visual and textual prompts, respectively.
The prompts are optimized with this objective during source training and fixed at target inference.

\subsection{Confidence and Context Features}
\label{sec:confidence_context}

CAF extracts instance-level confidence and collection-level contextual features from the four prediction branches (Fig.~\ref{fig:method_overview}(b)).
We describe these features for target inference; source training uses the same definitions with source-class probabilities and the historical EMA table introduced in Sec.~\ref{sec:optimization_inference}.
Let $C=|\mathcal C_t|$ and $p_k(c\mid x)=\operatorname{softmax}(z_k(x))_c$.
We characterize each branch's predictive uncertainty using normalized entropy:
\begin{equation}
    h_k(x)
    =-\frac{1}{\log C}
    \sum_{c\in\mathcal C_t}
    p_k(c\mid x)\log p_k(c\mid x).
    \label{eq:normalized_entropy}
\end{equation}
A lower entropy indicates a more concentrated prediction for the current image.

To interpret this confidence within the target context, we construct a shared soft prediction-frequency statistic for each unlabeled target collection:
\begin{equation}
    \bar p_t(c)
    =\frac{1}{4|\mathcal U_t|}
    \sum_{x\in\mathcal U_t}
    \sum_{k\in\mathcal K}
    p_k(c\mid x).
    \label{eq:shared_frequency}
\end{equation}
This statistic, denoted by $f(c)$ in Fig.~\ref{fig:method_overview}(b), aggregates class probabilities across all target images and all four branches, providing a common reference for their collective prediction tendencies.

For branch $k$, let $c_k(x)=\arg\max_{c\in\mathcal C_t}z_k(x,c)$ denote its top-ranked class.
We obtain the corresponding contextual feature through a negative-log-frequency transformation:
\begin{equation}
    r_k(x)
    =-\log\!\left(
    C\max\{\bar p_t(c_k(x)),\epsilon\}
    \right),
    \label{eq:context_feature}
\end{equation}
where $\epsilon>0$ ensures numerical stability.
The factor $C$ sets the uniform frequency $1/C$ as the reference: classes predicted less frequently receive larger contextual values.
Thus, $h_k(x)$ describes the concentration of an individual prediction, while $r_k(x)$ relates its proposed class to the collective prediction behavior.
These two features are subsequently combined to determine image-dependent fusion weights.

\subsection{Contextual Adaptive Prediction Fusion}
\label{sec:contextual_fusion}

CAF combines confidence and context to determine each branch's contribution (Fig.~\ref{fig:method_overview}(c)).
Let $d(x)=\mathbf 1\{\exists k\in\mathcal K:c_k(x)\ne c_{00}(x)\}$ indicate disagreement among the branch predictions.
We compute the fusion scores and weights as
\begin{equation}
    \begin{aligned}
        s_k(x)&=-\alpha h_k(x)+\beta d(x)r_k(x),\\
        w_k(x)&=\frac{\exp(s_k(x))}
        {\sum_{j\in\mathcal K}\exp(s_j(x))},
    \end{aligned}
    \label{eq:fusion_weights}
\end{equation}
where $\alpha$ and $\beta$ are learnable scalar coefficients balancing instance-level confidence and collection-level context.
The contextual term adjusts branch scores according to the collective frequency of their proposed classes when the branches disagree.

The weights are nonnegative, sum to one, and are shared across all classes for each image.
CAF uses them to fuse the original branch logits:
\begin{equation}
    \begin{aligned}
        z_{\mathrm{CAF}}(x,c)
        &=\sum_{k\in\mathcal K}w_k(x)z_k(x,c),\\
        \hat y(x)&=\arg\max_{c\in\mathcal C_t}z_{\mathrm{CAF}}(x,c).
    \end{aligned}
    \label{eq:fused_prediction}
\end{equation}
Thus, branch-proposed classes inform the weighting rule, while the final prediction considers the full target vocabulary.

We learn the fusion coefficients through cross-entropy on labeled source examples.
We first detach the branch logits, $\tilde z_k(x)=\operatorname{sg}(z_k(x))$, and compute both confidence and contextual features from these detached predictions.
The fusion objective is
\begin{equation}
    \mathcal L_{\mathrm{fuse}}
    =\frac{1}{|\mathcal B|}
    \sum_{(x,y)\in\mathcal B}
    \operatorname{CE}\!\left(
    \sum_{k\in\mathcal K}
    w_k(x;\alpha,\beta)\tilde z_k(x),\,y
    \right).
    \label{eq:fusion_objective}
\end{equation}
The overall training loss is
\begin{equation}
    \mathcal L_{\mathrm{total}}
    =\mathcal L_{\mathrm{prompt}}+\mathcal L_{\mathrm{fuse}}.
    \label{eq:total_objective}
\end{equation}
The stop-gradient operation ensures that $\mathcal L_{\mathrm{fuse}}$ updates only $\alpha$ and $\beta$, while $\mathcal L_{\mathrm{prompt}}$ optimizes the visual and textual prompts.
At inference, the learned coefficients remain fixed, while fusion weights vary across images according to their branch predictions and shared target statistics, without target labels or parameter updates.

\subsection{Optimization and Transductive Inference}
\label{sec:optimization_inference}

During source training, CAF maintains a shared exponential moving average (EMA) of predicted class probabilities over the source vocabulary.
Let $C_s=|\mathcal C_s|$ and initialize the table uniformly as $f^{(0)}(c)=1/C_s$.
For the augmented source minibatch $\mathcal B_n$ at iteration $n$, we compute
\begin{equation}
    \hat f^{(n)}(c)
    =\frac{1}{4|\mathcal B_n|}
    \sum_{(x,y)\in\mathcal B_n}
    \sum_{k\in\mathcal K}
    \operatorname{sg}\!\left(p_k(c\mid x)\right),
    \label{eq:batch_frequency}
\end{equation}
where branch probabilities are evaluated over $\mathcal C_s$.
This statistic averages predictions across all image views and branches, without using ground-truth class frequencies.

The current minibatch uses the existing table $f^{(n)}$ to compute contextual features and fusion weights.
Prompt parameters and fusion coefficients receive gradients from their respective objectives, with the branch outputs and frequency table detached from fusion learning.
After processing the minibatch, we update the table using its previously computed probabilities:
\begin{equation}
    f^{(n+1)}(c)
    =0.99f^{(n)}(c)+0.01\hat f^{(n)}(c).
    \label{eq:ema_update}
\end{equation}
Thus, each minibatch is weighted using historical statistics before its predictions enter the EMA.

At inference, all learned parameters are fixed, and the source EMA is replaced by target-specific statistics.
For each target collection $\mathcal U_t$, we first compute and cache the four branch logits while accumulating the shared probability average $\bar p_t$ over all unlabeled images.
We then traverse the cached logits to compute confidence and contextual features, derive image-dependent weights, and produce fused predictions.
The second pass requires no additional encoder evaluations.
Each target collection uses its own statistics, without target labels or parameter updates.

\begin{table*}[!t]
\centering
\caption{\textbf{Cross-dataset and cross-domain evaluation.} The backbone is ViT-B/16; accuracies are percentages. CAF results are mean $\pm$ sample standard deviation over three seeds, with standard deviations placed below the means in smaller type. The ImageNet column reports source accuracy. The two averages cover ten cross-dataset targets and four ImageNet variants, respectively. The best mean accuracy in each column is shown in bold.}
\label{zero-shot}
\fontsize{8}{9.5}\selectfont
\setlength{\tabcolsep}{0pt}
\renewcommand{\arraystretch}{1.12}
\newlength{\cafnumwidth}
\setlength{\cafnumwidth}{\dimexpr(\textwidth-80pt)/17\relax}
\newcommand{\cafstat}[2]{\makebox[\linewidth][c]{\vtop{\offinterlineskip\halign{\hfil##\hfil\cr#1\cr\noalign{\vskip1.2pt}{\fontsize{5.5}{6}\selectfont\textpm#2}\cr}}}}
\newcommand{\cafheader}[1]{\makebox[\linewidth][c]{{\fontsize{6.7}{7.6}\selectfont\renewcommand{\arraystretch}{1.0}\begin{tabular}[c]{@{}c@{}}#1\end{tabular}}}}
\begin{tabular}{@{}>{\raggedright\arraybackslash}p{80pt}*{17}{>{\centering\arraybackslash}p{\cafnumwidth}}@{}}
\toprule
&\multicolumn{1}{c}{\textbf{Source}}&\multicolumn{11}{c}{\textbf{Cross-dataset}}&\multicolumn{5}{c}{\textbf{Cross-domain}}\\
\cmidrule(lr){2-2}\cmidrule(lr){3-13}\cmidrule(lr){14-18}
Method & \cafheader{ImageNet\\(IN)} & \cafheader{Caltech\\101} & \cafheader{Oxford\\Pets} & \cafheader{Stanford\\Cars} & \cafheader{Flowers\\102} & \cafheader{Food\\101} & \cafheader{FGVC\\Aircraft} & \cafheader{SUN\\397} & \cafheader{DTD} & \cafheader{EuroSAT} & \cafheader{UCF\\101} & \cafheader{\textbf{Avg-10}} & \cafheader{IN-V2} & \cafheader{IN-S} & \cafheader{IN-A} & \cafheader{IN-R} & \cafheader{\textbf{Avg-4}} \\
\midrule
CoOp~\cite{zhou2022learning} & 71.51 & 93.70 & 89.10 & 64.50 & 68.70 & 85.30 & 18.50 & 64.20 & 41.90 & 46.40 & 66.60 & 63.90 & 64.20 & 47.99 & 49.71 & 75.21 & 59.28 \\
CoCoOp~\cite{zhou2022conditional} & 71.02 & 94.40 & 90.10 & 65.30 & 71.90 & 86.10 & 22.90 & 67.40 & 45.70 & 45.40 & 68.20 & 65.70 & 64.07 & 48.75 & 50.63 & 76.18 & 59.91 \\
TPT~\cite{shu2022test} & 68.98 & 94.10 & 87.70 & 66.80 & 68.90 & 84.60 & 24.70 & 65.50 & 47.70 & 42.40 & 68.00 & 65.10 & 63.45 & 47.94 & \textbf{54.77} & 77.06 & 60.81 \\
VPT~\cite{Derakhshani2022VariationalPT} & 70.70 & 93.70 & 90.60 & 65.00 & 70.90 & 86.30 & 24.90 & 67.50 & 46.10 & 45.90 & 68.70 & 66.00 & 64.23 & 49.20 & 51.33 & 77.00 & 60.44 \\
PromptSRC~\cite{khattak2023self} & 71.27 & 93.60 & 90.30 & 65.70 & 70.30 & 86.20 & 24.00 & 67.10 & 46.90 & 45.50 & 68.80 & 65.80 & 64.35 & 49.55 & 50.90 & 77.80 & 60.65 \\
CoPrompt~\cite{roy2024consistency} & 70.80 & 94.50 & 90.73 & 65.67 & 72.30 & 86.43 & 24.00 & 67.57 & 47.07 & 51.90 & 69.73 & 67.00 & 64.25 & 49.43 & 50.50 & 77.51 & 60.42 \\
ArGue~\cite{tian2024argue} & 71.57 & 94.30 & 88.90 & 67.20 & 69.80 & 85.00 & 24.40 & 66.80 & 46.20 & 46.10 & 69.20 & 65.80 & 64.57 & 48.92 & 50.93 & 76.56 & 60.25 \\
MMRL~\cite{guo2025mmrl} & 72.03 & 94.67 & 91.43 & 66.10 & 72.77 & 86.40 & 26.30 & 67.57 & 45.90 & 53.10 & 68.27 & 67.25 & 64.47 & 49.17 & 51.20 & 77.53 & 60.59 \\
TAC~\cite{hao2025task} & 72.77 & 94.53 & 90.67 & 65.30 & 72.20 & 85.83 & 23.53 & 67.63 & 47.57 & 48.07 & 70.00 & 66.53 & 65.97 & 50.30 & 51.73 & 78.50 & 61.63 \\
HiCroPL~\cite{zheng2025hierarchical} & 70.84 & 94.48 & 90.13 & 65.68 & 72.03 & 86.46 & 26.58 & 68.78 & 53.19 & 49.19 & 70.31 & 67.68 & 64.33 & 49.47 & 50.79 & 77.15 & 60.44 \\
SurPL-G~\cite{pmlr-v267-liu25bx} & \textbf{73.33} & 93.73 & 90.16 & 64.67 & 70.53 & 85.52 & 24.80 & 67.43 & 48.54 & 52.85 & 67.86 & 66.61 & \textbf{66.59} & 50.45 & 50.00 & 78.45 & 61.37 \\
MMRL++~\cite{guo2026mmrl} & 71.87 & 94.63 & 91.43 & 66.60 & \textbf{73.53} & 86.73 & 26.07 & 67.77 & 46.13 & 53.00 & 69.03 & 67.49 & 64.67 & 49.30 & 51.00 & 77.43 & 60.60 \\
AlignedNorm~\cite{ma2026alignednorm} & 71.83 & 94.50 & 91.57 & 66.50 & 73.00 & 86.67 & 26.30 & 67.97 & 46.63 & 53.47 & 69.50 & 67.61 & 65.13 & 49.53 & 51.10 & 78.07 & 60.96 \\
MMRL + ReBaPL~\cite{bendou2026rebapl} & 71.00 & 94.60 & 91.97 & 66.63 & 72.57 & 86.43 & 26.33 & 67.90 & 47.27 & 52.97 & 69.50 & 67.62 & 62.50 & 48.40 & 49.63 & 75.63 & 59.04 \\
MMRL + SpecPL~\cite{zhou2026specpl} & 72.07 & 94.85 & 91.99 & 66.48 & 72.96 & 86.86 & 26.58 & 67.94 & 46.43 & 52.88 & 69.10 & 67.61 & 64.60 & 49.33 & 51.43 & 77.67 & 60.76 \\
ManiPT~\cite{yang2026prompt} & 72.92 & \textbf{95.13} & 90.87 & 66.27 & 72.14 & 86.66 & 27.00 & 68.50 & 51.83 & 50.88 & 71.11 & 68.04 & 66.12 & 50.63 & 51.97 & \textbf{78.79} & \textbf{61.88} \\
\midrule
\rowcolor{revisionbg} \textbf{CAF (ours)} & \cafstat{72.26}{0.03} & \cafstat{94.10}{0.12} & \cafstat{\textbf{93.11}}{0.15} & \cafstat{\textbf{68.39}}{0.07} & \cafstat{71.88}{0.26} & \cafstat{\textbf{87.31}}{0.08} & \cafstat{\textbf{29.82}}{0.33} & \cafstat{\textbf{70.65}}{0.17} & \cafstat{\textbf{54.83}}{0.22} & \cafstat{\textbf{56.36}}{1.35} & \cafstat{\textbf{73.09}}{0.12} & \cafstat{\textbf{69.96}}{0.14} & \cafstat{65.63}{0.15} & \cafstat{\textbf{52.21}}{0.11} & \cafstat{49.63}{0.26} & \cafstat{77.97}{0.08} & \cafstat{61.36}{0.08} \\
\bottomrule
\end{tabular}
\end{table*}

\section{Experiments}
\label{sec:experiment}

\subsection{Experimental Setup}
\noindent\textbf{Datasets.} We consider eleven image-classification benchmarks: ImageNet~\cite{Deng2009ImageNetAL}, Caltech101~\cite{FeiFei2004LearningGV}, OxfordPets~\cite{parkhi2012cats}, StanfordCars~\cite{Krause20133DOR}, Flowers102~\cite{Nilsback2008AutomatedFC}, Food101~\cite{Bossard2014Food101M}, FGVCAircraft~\cite{Maji2013FineGrainedVC}, SUN397~\cite{Xiao2010SUNDL}, DTD~\cite{Cimpoi2014DescribingTI}, EuroSAT~\cite{Helber2019EuroSATAN}, and UCF101~\cite{Soomro2012UCF101AD}. Four additional ImageNet variants, ImageNetV2~\cite{Recht2019DoIC}, ImageNet-S~\cite{Wang2019LearningRG}, ImageNet-A, and ImageNet-R~\cite{Hendrycks2020TheMF}, provide distribution-shifted evaluation data.

\noindent\textbf{Implementation details.}
We freeze both CLIP ViT-B/16 encoders and train eight prompt tokens per layer with SGD and the fusion coefficients with Adam.
For cross-dataset and cross-domain transfer, we train on 16-shot ImageNet for 5 epochs.
For base-to-novel evaluation, we train separately on each dataset for 20 epochs using 16 labeled images per base class and compute fusion statistics separately on the unlabeled base and novel test splits.
Training hyperparameters are detailed in Sec.~\ref{app:implementation} of the supplement.

\begin{table*}[t!]
		\captionsetup{skip=4pt}
		\caption{\textbf{Base-to-novel generalization on 11 datasets.} Subscripts give sample standard deviations over three seeds, using seed-level HMs and dataset averages. $^{*}$ denotes dataset-specific loss weights. Bold and underline mark the best and second-best results.}
		\label{tab:base2novel_style_match}
		\renewcommand{\arraystretch}{0.84}
		\centering
		\scriptsize
		\newcommand{\bnostat}[2]{\mbox{#1\rlap{\kern.3pt\raisebox{-1.3pt}[0pt][0pt]{\fontsize{5}{5}\selectfont\textpm#2}}}}
		\setlength{\aboverulesep}{0.28ex}
		\setlength{\belowrulesep}{0.35ex}
		\setlength{\tabcolsep}{4pt}
			\newcolumntype{J}{>{\centering\arraybackslash}X<{\kern8pt\mbox{}}}
			\begin{tabularx}{\textwidth}{l JJJ|JJJ|JJJ|JJJ}
				\toprule
				\multirow{2}{*}{Method} & 
				\multicolumn{3}{c}{(a) Average} & 
				\multicolumn{3}{c}{(b) ImageNet} & 
				\multicolumn{3}{c}{(c) Caltech101} & 
				\multicolumn{3}{c}{(d) OxfordPets} \\
				\cmidrule(lr){2-4} \cmidrule(lr){5-7} \cmidrule(lr){8-10} \cmidrule(lr){11-13}
				& Base & Novel & HM & Base & Novel & HM & Base & Novel & HM & Base & Novel & HM \\
				\midrule
CLIP~\cite{radford2021learning} & 69.34 & 74.22 & 71.70 & 72.43 & 68.14 & 70.22 & 96.84 & 94.00 & 95.40 & 91.17 & 97.26 & 94.12 \\
CoOp~\cite{zhou2022learning} & 82.69 & 63.22 & 71.66 & 76.47 & 67.88 & 71.92 & 98.00 & 89.81 & 93.73 & 93.67 & 95.29 & 94.47 \\
CoCoOp~\cite{zhou2022conditional} & 80.47 & 71.69 & 75.83 & 75.98 & 70.43 & 73.10 & 97.96 & 93.81 & 95.84 & 95.20 & 97.69 & 96.43 \\
PromptSRC~\cite{khattak2023self} & 84.26 & 76.10 & 79.97 & 77.60 & 70.73 & 74.01 & 98.10 & 94.03 & 96.02 & 95.33 & 97.30 & 96.30 \\
CoPrompt~\cite{roy2024consistency} & 84.00 & 77.23 & 80.48 & 77.67 & 71.27 & 74.33 & 98.27 & 94.90 & 96.55 & 95.67 & 98.10 & 96.87 \\
TAP~\cite{ding2025tree} & 84.75 & 77.63 & 81.04 & 77.97 & 70.40 & 73.99 & 98.90 & 95.50 & 97.17 & 95.80 & 97.73 & 96.76 \\
TAC~\cite{hao2025task} & 85.24 & 77.60 & 81.24 & 78.57 & 71.03 & 74.61 & 98.57 & 95.27 & 96.89 & 95.93 & \textbf{98.17} & 97.04 \\
HiCroPL~\cite{zheng2025hierarchical} & 85.89 & 77.99 & 81.75 & 78.07 & 71.72 & 74.76 & 98.77 & \underline{95.96} & 97.34 & 96.28 & 97.76 & 97.01 \\
SurPL-G~\cite{pmlr-v267-liu25bx} & \textbf{86.37} & 76.32 & 81.03 & \textbf{78.74} & 70.49 & 74.39 & 98.77 & 95.16 & 96.93 & 96.37 & 97.41 & 96.89 \\
MMRL++~\cite{guo2026mmrl} & 85.53 & 78.32 & 81.77 & 77.63 & 71.50 & 74.44 & \underline{99.07} & 94.53 & 96.75 & 95.60 & 97.43 & 96.51 \\
MMRL + ReBaPL~\cite{bendou2026rebapl} & 85.74 & 77.44 & 81.38 & 77.90 & 68.83 & 73.09 & \textbf{99.10} & 94.27 & 96.62 & 95.93 & 97.57 & 96.74 \\
MMRL + SpecPL~\cite{zhou2026specpl} & 85.90 & 77.55 & 81.51 & 78.05 & 71.50 & 74.63 & \textbf{99.10} & 94.43 & 96.71 & 96.20 & 97.40 & 96.80 \\
ManiPT~\cite{yang2026prompt} & 85.82 & 78.67 & 82.09 & \underline{78.67} & 72.01 & 75.20 & 98.95 & 95.94 & \underline{97.42} & \underline{96.52} & \underline{98.15} & \underline{97.33} \\
\midrule
\rowcolor{revisionbg} \textbf{CAF} & \bnostat{85.78}{0.05} & \bnostat{\underline{79.52}}{0.50} & \bnostat{\underline{82.53}}{0.27} & \bnostat{78.36}{0.03} & \bnostat{\underline{73.15}}{0.15} & \bnostat{\underline{75.67}}{0.09} & \bnostat{98.77}{0.23} & \bnostat{95.89}{0.17} & \bnostat{97.31}{0.19} & \bnostat{96.10}{0.26} & \bnostat{98.02}{0.12} & \bnostat{97.05}{0.08} \\
\rowcolor{revisionbg} \textbf{CAF}$^{*}$ & \bnostat{\underline{86.09}}{0.15} & \bnostat{\textbf{79.77}}{0.23} & \bnostat{\textbf{82.81}}{0.19} & \bnostat{\underline{78.67}}{0.10} & \bnostat{\textbf{73.27}}{0.14} & \bnostat{\textbf{75.87}}{0.04} & \bnostat{98.75}{0.29} & \bnostat{\textbf{96.14}}{0.25} & \bnostat{\textbf{97.43}}{0.26} & \bnostat{\textbf{96.72}}{0.25} & \bnostat{97.97}{0.14} & \bnostat{\textbf{97.34}}{0.12} \\
				\midrule[\heavyrulewidth]
				\multirow{2}{*}{Method} & 
				\multicolumn{3}{c}{(e) StanfordCars} & 
				\multicolumn{3}{c}{(f) Flowers102} & 
				\multicolumn{3}{c}{(g) Food101} & 
				\multicolumn{3}{c}{(h) FGVCAircraft} \\
				\cmidrule(lr){2-4} \cmidrule(lr){5-7} \cmidrule(lr){8-10} \cmidrule(lr){11-13}
				& Base & Novel & HM & Base & Novel & HM & Base & Novel & HM & Base & Novel & HM \\
				\midrule
CLIP~\cite{radford2021learning} & 63.37 & 74.89 & 68.65 & 72.08 & \underline{77.80} & 74.83 & 90.10 & 91.22 & 90.66 & 27.19 & 36.29 & 31.09 \\
CoOp~\cite{zhou2022learning} & 78.12 & 60.40 & 68.13 & 97.60 & 59.67 & 74.06 & 88.33 & 82.26 & 85.19 & 40.44 & 22.30 & 28.75 \\
CoCoOp~\cite{zhou2022conditional} & 70.49 & 73.59 & 72.01 & 94.87 & 71.75 & 81.71 & 90.70 & 91.29 & 90.99 & 33.41 & 23.71 & 27.74 \\
PromptSRC~\cite{khattak2023self} & 78.27 & 74.97 & 76.58 & 98.07 & 76.50 & 85.95 & 90.67 & 91.53 & 91.10 & 42.73 & 37.87 & 40.15 \\
CoPrompt~\cite{roy2024consistency} & 76.97 & 74.40 & 75.66 & 97.27 & 76.60 & 85.71 & 90.73 & 92.07 & 91.40 & 40.20 & 39.33 & 39.76 \\
TAP~\cite{ding2025tree} & 80.70 & 74.27 & 77.35 & 97.90 & 75.57 & 85.30 & 90.97 & 91.83 & 91.40 & 44.40 & 36.50 & 40.06 \\
TAC~\cite{hao2025task} & 81.63 & 74.17 & 77.72 & 97.97 & 76.87 & 86.15 & 90.87 & 91.87 & 91.37 & 44.60 & 37.70 & 40.86 \\
HiCroPL~\cite{zheng2025hierarchical} & 81.51 & 75.04 & 78.14 & 98.29 & 75.46 & 85.38 & 90.96 & 91.67 & 91.31 & \underline{48.38} & 41.75 & \underline{44.82} \\
SurPL-G~\cite{pmlr-v267-liu25bx} & \underline{83.57} & 72.77 & 77.80 & \textbf{98.90} & 72.88 & 83.92 & 90.92 & 91.81 & 91.36 & \textbf{49.20} & 36.93 & 42.19 \\
MMRL++~\cite{guo2026mmrl} & 81.33 & 75.27 & 78.18 & 98.53 & \textbf{77.90} & \textbf{87.01} & 90.47 & 91.73 & 91.10 & 46.40 & 38.77 & 42.24 \\
MMRL + ReBaPL~\cite{bendou2026rebapl} & 81.20 & 75.37 & 78.17 & \underline{98.80} & 77.23 & \underline{86.70} & 90.73 & 91.60 & 91.16 & 45.13 & 38.57 & 41.59 \\
MMRL + SpecPL~\cite{zhou2026specpl} & 82.20 & 75.50 & 78.71 & \underline{98.80} & 77.20 & 86.67 & 91.00 & 91.70 & 91.35 & 46.20 & 37.80 & 41.58 \\
ManiPT~\cite{yang2026prompt} & 82.53 & 75.61 & 78.92 & 98.12 & 76.84 & 86.19 & \textbf{91.17} & 92.16 & \underline{91.66} & 48.28 & 39.87 & 43.68 \\
\midrule
\rowcolor{revisionbg} \textbf{CAF} & \bnostat{82.95}{0.57} & \bnostat{\textbf{76.96}}{0.30} & \bnostat{\underline{79.84}}{0.30} & \bnostat{98.16}{0.24} & \bnostat{76.83}{0.50} & \bnostat{86.20}{0.37} & \bnostat{90.45}{0.23} & \bnostat{\textbf{92.53}}{0.02} & \bnostat{91.48}{0.13} & \bnostat{45.70}{0.44} & \bnostat{\underline{43.01}}{2.70} & \bnostat{44.31}{1.40} \\
\rowcolor{revisionbg} \textbf{CAF}$^{*}$ & \bnostat{\textbf{83.70}}{0.63} & \bnostat{\underline{76.76}}{0.33} & \bnostat{\textbf{80.08}}{0.41} & \bnostat{98.16}{0.24} & \bnostat{76.83}{0.50} & \bnostat{86.20}{0.37} & \bnostat{\underline{91.02}}{0.17} & \bnostat{\underline{92.50}}{0.08} & \bnostat{\textbf{91.75}}{0.13} & \bnostat{46.18}{1.08} & \bnostat{\textbf{44.63}}{0.91} & \bnostat{\textbf{45.39}}{0.59} \\
				\midrule[\heavyrulewidth]
				\multirow{2}{*}{Method} & 
				\multicolumn{3}{c}{(i) SUN397} & 
				\multicolumn{3}{c}{(j) DTD} & 
				\multicolumn{3}{c}{(k) EuroSAT} & 
				\multicolumn{3}{c}{(l) UCF101} \\
				\cmidrule(lr){2-4} \cmidrule(lr){5-7} \cmidrule(lr){8-10} \cmidrule(lr){11-13}
				& Base & Novel & HM & Base & Novel & HM & Base & Novel & HM & Base & Novel & HM \\
				\midrule
CLIP~\cite{radford2021learning} & 69.36 & 75.35 & 72.23 & 53.24 & 59.90 & 56.37 & 56.48 & 64.05 & 60.03 & 70.53 & 77.50 & 73.85 \\
CoOp~\cite{zhou2022learning} & 80.60 & 65.89 & 72.51 & 79.44 & 41.18 & 54.24 & 92.19 & 54.74 & 68.69 & 84.69 & 56.05 & 67.46 \\
CoCoOp~\cite{zhou2022conditional} & 79.74 & 76.86 & 78.27 & 77.01 & 56.00 & 64.85 & 87.49 & 60.04 & 71.21 & 82.33 & 73.45 & 77.64 \\
PromptSRC~\cite{khattak2023self} & 82.67 & 78.47 & 80.52 & 83.37 & 62.97 & 71.75 & 92.90 & 73.90 & 82.32 & 87.10 & 78.80 & 82.74 \\
CoPrompt~\cite{roy2024consistency} & 82.63 & 80.03 & 81.31 & 83.13 & 64.73 & 72.79 & 94.60 & 78.57 & 85.84 & 86.90 & 79.57 & 83.07 \\
TAP~\cite{ding2025tree} & 82.87 & 79.53 & 81.17 & 84.20 & 68.00 & 75.24 & 90.70 & 82.17 & 86.22 & 87.90 & 82.43 & 85.08 \\
TAC~\cite{hao2025task} & \textbf{83.70} & 80.03 & 81.82 & 83.37 & 64.27 & 72.58 & 94.37 & 82.60 & 88.10 & 88.07 & 81.67 & 84.75 \\
HiCroPL~\cite{zheng2025hierarchical} & 83.23 & 79.92 & 81.54 & 85.07 & 67.34 & 75.17 & 96.29 & 80.36 & 87.61 & 87.95 & 80.91 & 84.28 \\
SurPL-G~\cite{pmlr-v267-liu25bx} & 83.43 & 78.96 & 81.13 & \underline{86.07} & 62.04 & 72.11 & 94.63 & 81.33 & 87.48 & \textbf{89.44} & 79.74 & 84.31 \\
MMRL++~\cite{guo2026mmrl} & 83.03 & 79.60 & 81.28 & 85.47 & 65.97 & 74.46 & 95.93 & \textbf{88.27} & \textbf{91.94} & 87.37 & 80.53 & 83.81 \\
MMRL + ReBaPL~\cite{bendou2026rebapl} & 83.20 & 79.37 & 81.24 & \textbf{86.10} & 65.00 & 74.08 & \underline{96.73} & 83.63 & \underline{89.71} & 88.33 & 80.40 & 84.18 \\
MMRL + SpecPL~\cite{zhou2026specpl} & \underline{83.57} & 79.11 & 81.28 & 85.52 & 65.72 & 74.32 & 95.40 & 81.45 & 87.87 & \underline{88.85} & 81.20 & 84.85 \\
ManiPT~\cite{yang2026prompt} & 83.18 & 80.18 & 81.65 & 84.38 & 68.56 & 75.65 & 94.06 & \underline{83.72} & 88.59 & 88.14 & 82.31 & 85.13 \\
\midrule
\rowcolor{revisionbg} \textbf{CAF} & \bnostat{83.36}{0.16} & \bnostat{\underline{81.35}}{0.20} & \bnostat{\underline{82.34}}{0.04} & \bnostat{85.30}{0.87} & \bnostat{\underline{69.97}}{0.77} & \bnostat{\underline{76.88}}{0.78} & \bnostat{96.13}{0.25} & \bnostat{83.61}{2.13} & \bnostat{89.43}{1.20} & \bnostat{88.26}{0.49} & \bnostat{\textbf{83.36}}{0.68} & \bnostat{\underline{85.74}}{0.19} \\
\rowcolor{revisionbg} \textbf{CAF}$^{*}$ & \bnostat{83.41}{0.29} & \bnostat{\textbf{81.70}}{0.18} & \bnostat{\textbf{82.55}}{0.06} & \bnostat{84.53}{0.44} & \bnostat{\textbf{71.26}}{1.60} & \bnostat{\textbf{77.33}}{1.03} & \bnostat{\textbf{97.14}}{0.62} & \bnostat{83.19}{0.58} & \bnostat{89.63}{0.41} & \bnostat{88.69}{0.31} & \bnostat{\underline{83.22}}{1.37} & \bnostat{\textbf{85.87}}{0.68} \\
				\bottomrule
			\end{tabularx}
		
\end{table*}

\subsection{Cross-Dataset and Cross-Domain Generalization}
As shown in Tab.~\ref{zero-shot}, CAF achieves the highest average accuracy of $69.96\%\pm0.14\%$ across the ten cross-dataset benchmarks, surpassing the strongest compared baseline, ManiPT~\cite{yang2026prompt}, by 1.92 percentage points.
It also outperforms the recent MMRL-based ReBaPL~\cite{bendou2026rebapl} and SpecPL~\cite{zhou2026specpl} variants by 2.34 and 2.35 percentage points, respectively.
CAF ranks first on eight of the ten datasets, spanning fine-grained recognition~\cite{Maji2013FineGrainedVC}, scene classification~\cite{Xiao2010SUNDL}, texture recognition~\cite{Cimpoi2014DescribingTI}, remote sensing~\cite{Helber2019EuroSATAN}, and action recognition~\cite{Soomro2012UCF101AD}, indicating broad transferability across diverse visual domains and class vocabularies.
These target improvements are achieved despite lower ImageNet accuracy than ManiPT and TAC~\cite{hao2025task}, showing that higher source accuracy does not necessarily translate into stronger cross-dataset generalization.
CAF also maintains competitive robustness on the four ImageNet variants, achieving an average accuracy of $61.36\%\pm0.08\%$, exceeding 13 of the 16 compared baselines, and obtaining the best result among them on ImageNet-S at 52.21\%.
These results indicate that CAF mitigates overfitting to the source class vocabulary, delivering substantial advantages under joint visual and semantic shifts while maintaining competitive robustness across domain shifts.

\subsection{Base-to-Novel Generalization}
With a single loss-weight pair used for all eleven datasets, CAF attains 79.52\% average novel-class accuracy and a harmonic mean (HM) of 82.53\%, outperforming every baseline on both metrics, while its average base accuracy (85.78\%) is within 0.04 percentage points of ManiPT~\cite{yang2026prompt}, the strongest baseline on novel accuracy and HM in Tab.~\ref{tab:base2novel_style_match}.
This 0.85-point novel gain, with base accuracy essentially unchanged (0.04 points lower), raises HM by 0.44 points and yields the smallest base--novel gap among the prompt-learning methods compared (6.26 vs.\ 7.15 points for ManiPT).
Compared with the baseline with the highest base accuracy (SurPL-G~\cite{pmlr-v267-liu25bx}, 86.37\%), CAF incurs a 0.59-point drop in base accuracy but gains 3.20 points on novel classes and 1.50 points on HM, indicating a more favorable base--novel trade-off.

CAF$^{*}$, which uses dataset-specific loss weights, further raises base accuracy, novel accuracy, and HM to 86.09\%, 79.77\%, and 82.81\%, exceeding ManiPT on all three metrics (+0.27 base, +1.10 novel, and +0.72 HM).
CAF$^{*}$ achieves the best HM on nine of the eleven datasets and matches or exceeds ManiPT's HM on every dataset.
Its largest novel-class gains over the best baseline result on each dataset are 2.88 and 2.70 points on FGVCAircraft (44.63\%) and DTD (71.26\%). These are also the two benchmarks with the lowest best-baseline novel accuracy.
Despite this per-dataset tuning, the average gain over CAF is at most 0.31 points on any metric. The shared setting thus captures most of the benefit of loss-weight selection, and the improvement over baselines does not hinge on dataset-specific tuning.

\subsection{Ablation Study}

\noindent\textbf{Fusion weights.}
Tab.~\ref{tab:ablation_study}\subref{tab:fusion_ablation} shows that CAF improves generalization beyond both single-branch prediction ($\mathbf e_{VT}$) and static fusion.
Uniform fusion ($\frac{1}{4}\mathbf 1$) outperforms both source-fitted ($\mathbf w_{\mathrm{src}}$) and learned image-independent weights ($\mathbf w_{\theta}$) on the target benchmarks, suggesting that optimizing a global branch preference on the source domain does not necessarily improve transfer.
CAF further exceeds uniform fusion by 1.89 and 0.48 percentage points on Avg-10 and Avg-4, respectively, while also improving ImageNet accuracy.
These results support adapting branch contributions to individual images rather than applying the same weights to every input.

\begin{table*}[t!]
    \centering
    \caption{\textbf{CAF ablations.} Accuracies (\%) on ImageNet (IN), ten cross-dataset targets (Avg-10), and four cross-domain targets (Avg-4). Time denotes median end-to-end inference time in seconds; Rel. is its ratio to the confidence-only baseline. $\uparrow$/$\downarrow$ indicate that higher/lower is better.}
    \label{tab:ablation_study}
    \captionsetup[subtable]{position=top,font=footnotesize,skip=4pt,justification=centering,singlelinecheck=false}
    \begin{subtable}[t]{0.28\textwidth}
        \centering
        \caption{Fusion weights}
        \label{tab:fusion_ablation}
        \footnotesize
        \setlength{\tabcolsep}{1.4pt}
        \renewcommand{\arraystretch}{1.16}
        \begin{tabular*}{\linewidth}{@{\extracolsep{\fill}}lrrr@{}}
            \toprule
            Weights & IN$\uparrow$ & Avg-10$\uparrow$ & Avg-4$\uparrow$ \\
            \midrule
            $\mathbf e_{VT}$ & 72.624 & 65.426 & 60.589 \\
            $\mathbf w_{\mathrm{src}}$ & 72.590 & 65.812 & 60.671 \\
            $\mathbf w_{\theta}$ & 72.316 & 66.241 & 60.611 \\
            $\frac{1}{4}\mathbf 1$ & 72.058 & 67.638 & 61.169 \\
            $\mathbf w(x)$ (\textbf{CAF}) & \textbf{72.724} & \textbf{69.524} & \textbf{61.649} \\
            \bottomrule
        \end{tabular*}
    \end{subtable}\hfill
    \begin{subtable}[t]{0.415\textwidth}
        \centering
        \caption{Confidence and context}
        \label{tab:frequency_statistics}
        \footnotesize
        \setlength{\tabcolsep}{1.4pt}
        \renewcommand{\arraystretch}{1.16}
        \begin{tabular*}{\linewidth}{@{\extracolsep{\fill}}lrrrr@{}}
            \toprule
            Inputs $(h_k,r_k)$ & Avg-10$\uparrow$ & Avg-4$\uparrow$ & Time (s)$\downarrow$ & Rel.$\downarrow$ \\
            \midrule
            $(h_k,0)$ & 68.122 & 61.106 & 111.876 & 1.0000$\times$ \\
            $(0,r_k^{\mathrm{br}})$ & 69.236 & 61.444 & \textbf{111.776} & \textbf{0.9991}$\times$ \\
            $(h_k,r_k^{\mathrm{br}})$ & 69.405 & 61.641 & 112.592 & 1.0064$\times$ \\
            $(h_k,r_k^{\mathrm{hard}})$ & 69.237 & 61.465 & 112.437 & 1.0050$\times$ \\
            $(h_k,r_k^{\mathrm{sh}})$ (\textbf{CAF}) & \textbf{69.524} & \textbf{61.649} & 112.560 & 1.0061$\times$ \\
            \bottomrule
        \end{tabular*}
    \end{subtable}\hfill
    \begin{subtable}[t]{0.28\textwidth}
        \centering
        \caption{Prompt regularization}
        \label{tab:regularization_ablation}
        \footnotesize
        \setlength{\tabcolsep}{1.4pt}
        \renewcommand{\arraystretch}{1.16}
        \begin{tabular*}{\linewidth}{@{\extracolsep{\fill}}ccrrr@{}}
            \toprule
            $\mathcal L_{\mathrm{margin}}$ & $\mathcal L_{\mathrm{text}}$ & IN$\uparrow$ & Avg-10$\uparrow$ & Avg-4$\uparrow$ \\
            \midrule
            -- & -- & \textbf{73.216} & 69.043 & 61.353 \\[3.67pt]
            $\checkmark$ & -- & 72.958 & 68.745 & \textbf{61.484} \\[3.67pt]
            -- & $\checkmark$ & 72.510 & 69.845 & 61.103 \\[3.67pt]
            $\checkmark$ & $\checkmark$ & 72.263 & \textbf{69.955} & 61.361 \\
            \bottomrule
        \end{tabular*}
    \end{subtable}
\end{table*}

\begin{figure*}[!t]
    \centering
    \includegraphics[width=\textwidth]{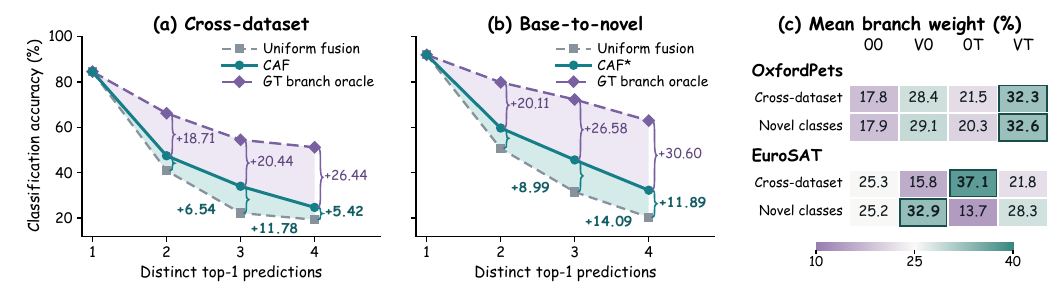}
    \caption{\textbf{Adaptive fusion across transfer settings.} (a,b) Accuracy by branch disagreement across ten datasets. Braces give differences in percentage points. (c) Mean branch weights on OxfordPets and EuroSAT. Outlines mark the largest weight in each row.}
    \label{fig:branch_disagreement}
\end{figure*}

\noindent\textbf{Confidence and context.}
Tab.~\ref{tab:ablation_study}\subref{tab:frequency_statistics} shows the benefit of incorporating collection-level context into prediction fusion.
CAF improves over confidence-only fusion by 1.40 and 0.54 percentage points on Avg-10 and Avg-4, respectively.
With branch-specific soft frequencies ($r_k^{\mathrm{br}}$), combining confidence and context also outperforms either signal alone, supporting their complementary roles.
Among the evaluated context estimators, shared soft frequencies ($r_k^{\mathrm{sh}}$) achieve the highest average accuracies: they improve Avg-10 over branch-specific statistics while maintaining comparable Avg-4, and outperform shared hard frequencies ($r_k^{\mathrm{hard}}$) on both metrics.
The measured end-to-end inference overhead is only 0.61\% relative to confidence-only fusion, indicating that the contextual information can be incorporated at little additional computational cost.

\noindent\textbf{Prompt regularization.}
Tab.~\ref{tab:ablation_study}\subref{tab:regularization_ablation} reveals different generalization trade-offs for the two regularizers.
The margin-only and text-only configurations are selected by Avg-4 and Avg-10, respectively.
Text regularization alone increases Avg-10 from 69.043\% to 69.845\%, whereas margin regularization alone achieves the highest Avg-4 of 61.484\% among the reported configurations.
The combined setting obtains the highest Avg-10 of 69.955\% and improves Avg-4 over text regularization alone, although it remains below the margin-only result.
Thus, the selected combination favors cross-dataset transfer while recovering part of the cross-domain performance lost with text regularization alone.
The complete coefficient sweep is provided in Fig.~\ref{fig:regularization_sweep}.

\subsection{Analysis}
\label{sec:analysis_experiments}

\noindent\textbf{Branch disagreement.}
Fig.~\ref{fig:branch_disagreement}(a,b) groups test images by the number of distinct top-1 predictions among the four branches.
When all branches agree, adaptive and uniform fusion produce the same prediction.
At every disagreement level under both protocols, adaptive fusion outperforms uniform fusion, with the largest gain when three distinct classes are proposed ($+11.8$ and $+14.1$ percentage points for CAF in cross-dataset transfer and CAF$^{*}$ on novel classes, respectively); the gain remains positive even when all four predictions differ.
A ground-truth branch oracle that selects a correct branch whenever one exists remains 19--31 points above adaptive fusion across disagreement levels, indicating that the branches already contain complementary correct predictions and that estimating which branch to trust remains the main bottleneck.

\noindent\textbf{Branch preferences.}
Fig.~\ref{fig:branch_disagreement}(c) compares mean fusion weights on a natural-object benchmark (OxfordPets) and a satellite-imagery benchmark (EuroSAT).
On OxfordPets, the fully prompted $VT$ branch receives the largest weight under both protocols.
On EuroSAT, the dominant branch changes across protocols: with ImageNet-trained prompts, $0T$ (frozen visual features with prompted text) receives the largest weight, whereas after base-class training, $V0$ (prompted visual features with frozen text) leads on novel classes.
The preferred branch is therefore not a fixed property of the method but depends on the transfer setting, which is why a single fixed branch or global weighting rule cannot be optimal across settings.
Sec.~\ref{app:mean_fusion_weights} reports the full comparison across datasets.

\section{Conclusion}
\label{sec:conclusion}

Our study suggests a clear answer to when to trust adaptation, namely neither globally nor uniformly, but locally and contextually. Contextual Adaptive Fusion (CAF) treats frozen and prompted visual and textual representations as four complementary branches, weighting them on each image by combining instance-level branch confidence with collection-level prediction frequency. With fusion coefficients fixed after source training, target-time arbitration relies solely on unlabeled target evidence, elevating the frozen model from a training-time regularizer to an active inference-time evidence source.
This perspective explains why a single static balance is insufficient, as adaptive gains concentrate on images with branch disagreement while dominant branch preferences shift across transfer regimes even within the same domain. Consequently, CAF achieves SOTA cross-dataset transfer, strong robustness under ImageNet distribution shifts, and superior base-to-novel generalization without per-target tuning. The remaining oracle headroom further confirms that the four branches already contain the evidence needed for harder cases, making richer branch interactions and online context estimation promising directions for future work.

\endgroup

\section*{Acknowledgments}
This research was supported by the Fund of the National Natural Science Foundation of China (Grant No.~62576177), Shenzhen Science and Technology Program (QNXMB20250701090801002, JCYJ20250604184027034, JCYJ20240813114237048), Guangdong Basic and Applied Basic Research Foundation (2026A1515011435), the Fundamental Research Funds for the Central Universities (070-63263247), Youth Interdisciplinary Research Project of Nankai University (Grant No.~NKQNJC202620), and the Tianjin Key Laboratory of Visual Computing and Intelligent Perception (VCIP).
Additional support was provided by the ``Science and Technology Yongjiang 2035'' key technology breakthrough plan project (2025Z053), Chinese government-guided local science and technology development fund projects (scientific and technological achievement transfer and transformation projects) (254Z0102G), and the National Science Fund of China (Grant No.~62361166670).
Computation is supported by the Supercomputing Center of Nankai University (NKSC).

\section*{LLM Usage Statement}
OpenAI Codex was used to assist with manuscript drafting and language revision, \LaTeX{} formatting, preparation of analysis figures and literature-comparison tables, and consistency checks of references and reported results.
The authors verified all citations, numerical results, scientific claims, and final content and take full responsibility for the manuscript.

{
    \small
    \bibliographystyle{ieeenat_fullname}
    \bibliography{main}
}

\clearpage
\appendix
\setcounter{section}{0}
\setcounter{subsection}{0}
\setcounter{equation}{0}
\setcounter{table}{0}
\setcounter{figure}{0}
\newcommand{\supplementprefix}{A}
\renewcommand{\theequation}{\supplementprefix\arabic{equation}}
\renewcommand{\thetable}{\supplementprefix\arabic{table}}
\renewcommand{\thefigure}{\supplementprefix\arabic{figure}}
\renewcommand{\theHequation}{appendix.\supplementprefix.\arabic{equation}}
\renewcommand{\theHtable}{appendix.\supplementprefix.\arabic{table}}
\renewcommand{\theHfigure}{appendix.\supplementprefix.\arabic{figure}}
\newcommand{\nextsupplementpart}[1]{%
    \renewcommand{\supplementprefix}{#1}%
    \setcounter{equation}{0}%
    \setcounter{table}{0}%
    \setcounter{figure}{0}%
}

\twocolumn[{
\centering
\Large\bfseries Supplementary Material\par
\vspace{1em}
}]

\begin{table}[H]
\centering
\caption{Dataset statistics. Training counts exclude validation images and precede few-shot sampling and base/novel partitioning. A dash indicates a target-only dataset.}
\label{tab:dataset_statistics}
\fontsize{8.5}{10}\selectfont
\setlength{\tabcolsep}{2.5pt}
\renewcommand{\arraystretch}{1.08}
\begin{tabular*}{\columnwidth}{@{\extracolsep{\fill}}llrrr@{}}
\toprule
Dataset & Task / domain & Classes & Train & Test \\
\midrule
\multicolumn{5}{l}{\textit{Standard benchmarks}}\\
ImageNet~\cite{Deng2009ImageNetAL} & Objects & 1,000 & 1,281,167 & 50,000 \\
Caltech101~\cite{FeiFei2004LearningGV} & Objects & 100 & 4,128 & 2,465 \\
OxfordPets~\cite{parkhi2012cats} & Fine-grained & 37 & 2,944 & 3,669 \\
StanfordCars~\cite{Krause20133DOR} & Fine-grained & 196 & 6,509 & 8,041 \\
Flowers102~\cite{Nilsback2008AutomatedFC} & Fine-grained & 102 & 4,093 & 2,463 \\
Food101~\cite{Bossard2014Food101M} & Fine-grained & 101 & 50,500 & 30,300 \\
FGVCAircraft~\cite{Maji2013FineGrainedVC} & Fine-grained & 100 & 3,334 & 3,333 \\
SUN397~\cite{Xiao2010SUNDL} & Scenes & 397 & 15,880 & 19,850 \\
DTD~\cite{Cimpoi2014DescribingTI} & Textures & 47 & 2,820 & 1,692 \\
EuroSAT~\cite{Helber2019EuroSATAN} & Remote sensing & 10 & 13,500 & 8,100 \\
UCF101~\cite{Soomro2012UCF101AD} & Actions & 101 & 7,639 & 3,783 \\
\midrule
\multicolumn{5}{l}{\textit{ImageNet variants}}\\
ImageNetV2~\cite{Recht2019DoIC} & Natural shift & 1,000 & -- & 10,000 \\
ImageNet-Sketch~\cite{Wang2019LearningRG} & Sketches & 1,000 & -- & 50,889 \\
ImageNet-A & Adversarial & 200 & -- & 7,500 \\
ImageNet-R~\cite{Hendrycks2020TheMF} & Renditions & 200 & -- & 30,000 \\
\bottomrule
\end{tabular*}
\end{table}

\section{Datasets and Experimental Details}
\label{app:supplementary}

\subsection{Datasets and Evaluation Protocols}
\label{app:dataset_details}

We consider eleven standard image-classification benchmarks and four distribution-shifted ImageNet variants, summarized in Tab.~\ref{tab:dataset_statistics}.
These datasets cover general object recognition, fine-grained recognition, scene classification, texture recognition, remote sensing, and action recognition.
We use RGB images for EuroSAT and extracted video frames for UCF101.

For cross-dataset evaluation, ImageNet provides the labeled source data, and the remaining ten standard datasets serve as targets.
Cross-domain evaluation uses ImageNetV2, ImageNet-Sketch, ImageNet-A, and ImageNet-R.
ImageNetV2 and ImageNet-Sketch retain the 1,000 ImageNet classes, whereas ImageNet-A and ImageNet-R contain subsets of 200 classes.
ImageNet-S in Tab.~\ref{zero-shot} denotes ImageNet-Sketch.
Avg-10 and Avg-4 average accuracies over the ten cross-dataset and four cross-domain targets, respectively, excluding source ImageNet accuracy.

For base-to-novel evaluation, each of the eleven standard datasets is partitioned into disjoint base and novel class sets.
Prompts are trained using labeled images from base classes and evaluated separately on base and novel classes, with novel classes unseen during training.
We report base accuracy, novel accuracy, and their harmonic mean (HM) to assess the balance between adaptation and generalization.

\subsection{Implementation Details}
\label{app:implementation}

\begin{table}[!t]
    \centering
    \caption{Dataset-specific loss weights used by CAF$^{*}$ for base-to-novel generalization in Tab.~\ref{tab:base2novel_style_match}.}
    \label{tab:base2novel_loss_weights}
    \small
    \begin{tabular*}{\columnwidth}{@{\extracolsep{\fill}}lrr@{}}
        \toprule
        Dataset(s) & $\lambda_{\mathrm{margin}}$ & $\lambda_{\mathrm{text}}$ \\
        \midrule
        ImageNet, EuroSAT & 1 & 2.5 \\
        Caltech101 & 0.5 & 0 \\
        OxfordPets & 2 & 20 \\
        StanfordCars, UCF101 & 0 & 2.5 \\
        Flowers102 & 4 & 2.5 \\
        Food101 & 4 & 40 \\
        FGVCAircraft & 2 & 0 \\
        SUN397 & 2 & 5 \\
        DTD & 4 & 5 \\
        \bottomrule
    \end{tabular*}
\end{table}

\noindent\textbf{Backbone and prompts.}
We use CLIP ViT-B/16 with eight learnable prompt tokens at each of the 12 layers in both the visual and text encoders.
The pretrained encoders and logit scale remain frozen.
Following ManiPT and CoPrompt~\cite{yang2026prompt,roy2024consistency}, we construct frozen textual prototypes from fixed GPT-generated class descriptions.
Throughout this section, loss-weight pairs are given in the order $(\lambda_{\mathrm{margin}},\lambda_{\mathrm{text}})$.

\noindent\textbf{Source training.}
For cross-dataset and cross-domain evaluation, we train on 16-shot ImageNet for 5 epochs.
Each minibatch contains 32 source images with four augmented views per image at a resolution of $224\times224$.
We optimize the prompts with SGD at a peak learning rate of $0.01$ and the fusion coefficients $\alpha,\beta$ with Adam at a peak learning rate of $10^{-3}$, using a one-epoch warm-up followed by cosine decay.
We use the same loss weights $(1,40)$ and the same trained model for all ten cross-dataset targets and four ImageNet variants.
The fusion loss has unit weight.
During training, we maintain a shared exponential moving average of branch prediction probabilities with a decay of $0.99$.
At inference, we fix all learned parameters and recompute the shared statistics from the full unlabeled test set of each target dataset.
Tab.~\ref{zero-shot} reports mean accuracies and sample standard deviations over three seeds.

\noindent\textbf{Base-to-novel training.}
We train a separate model on each dataset using a fixed base/novel class split and 16 labeled images per base class.
On the ten downstream datasets, we train for 20 epochs using minibatches of four images with four augmented views each (16 views per update) and a peak prompt learning rate of $0.0025$.
On ImageNet, the split contains 500 base and 500 novel classes, giving 8,000 base training images.
We use 32 images per minibatch with four views each (128 views per update) and a peak prompt learning rate of $0.01$.
The 10- and 20-epoch ImageNet runs are trained independently from scratch.
Tab.~\ref{tab:base2novel_style_match} uses the final checkpoint of the 20-epoch run for every dataset.
Across all eleven datasets, prompt SGD uses momentum $0.9$, weight decay $5\times10^{-4}$, and a first-epoch learning rate of $10^{-5}$ followed by cosine decay from the dataset-specific peak.
We retain the Adam optimizer and unit fusion-loss weight used in source training.
At evaluation, we use batches of 64 images and compute the shared statistics separately over the full unlabeled base and novel test splits.

For CAF, the shared loss weights $(4,2.5)$ give the highest harmonic mean of the ten-dataset average base and novel accuracies among the 42 configurations in Fig.~\ref{fig:b2n_regularization_sweep}.
We use this pair on all eleven datasets, including ImageNet.
For CAF$^{*}$, we allow the loss weights to vary by dataset and select the combination that maximizes the same harmonic mean over the ten downstream datasets.
On ImageNet, we select $(1,2.5)$ by fused base accuracy among the three configurations evaluated for 20 epochs.
Tab.~\ref{tab:base2novel_loss_weights} lists the resulting dataset-specific settings.

For both CAF variants, we use fusion on novel classes for all datasets.
On base classes, we use fusion on ImageNet and the $VT$ branch on the other ten datasets.
Tab.~\ref{tab:base2novel_style_match} reports three-seed mean base and novel accuracies and computes HM from these means.

\nextsupplementpart{B}
\section{Additional Ablation Studies}
\label{app:additional_ablations}


\begingroup
\setlength{\textfloatsep}{12pt plus 1pt minus 2pt}
\captionsetup[table]{skip=4pt}
\begin{table}[!t]
    \centering
    \caption{\textbf{Individual branches and prediction fusion.} Branch and uniform-fusion accuracies (\%) share one historical checkpoint; CAF repeats the learned-coefficient, three-seed means from Tab.~\ref{zero-shot}.}
    \label{tab:individual_branches}
    \small
    \setlength{\tabcolsep}{3pt}
    \renewcommand{\arraystretch}{1.02}
    \begin{tabular*}{\columnwidth}{@{\extracolsep{\fill}}lrrr@{}}
        \toprule
        Prediction & ImageNet$\uparrow$ & Avg-10$\uparrow$ & Avg-4$\uparrow$ \\
        \midrule
        \texttt{00} & 69.642 & 66.873 & 60.024 \\
        \texttt{V0} & 70.882 & 66.272 & 59.871 \\
        \texttt{0T} & 71.220 & 66.766 & 60.762 \\
        \texttt{VT} & 72.184 & 65.884 & 60.487 \\
        \midrule
        Uniform fusion & 71.616 & 67.517 & 60.962 \\
        \textbf{CAF (ours)} & \textbf{72.263} & \textbf{69.955} & \textbf{61.361} \\
        \bottomrule
    \end{tabular*}
\end{table}

\subsection{Individual Branches and Fusion}
\label{app:individual_branches}

Tab.~\ref{tab:individual_branches} compares the four individual prediction branches with uniform fusion using cached logits from the same prompt checkpoint.
Among the individual branches, $VT$ achieves the highest ImageNet accuracy but the lowest Avg-10 accuracy, indicating that stronger source performance does not necessarily translate into better cross-dataset generalization.
The frozen $00$ branch achieves the highest Avg-10, whereas $0T$ performs best on Avg-4, revealing different strengths across the evaluation settings.

Uniform fusion exceeds every individual branch on both target averages, reaching 67.517\% on Avg-10 and 60.962\% on Avg-4.
These results support the complementarity of frozen and prompted predictions and motivate retaining multiple branches for transfer.
The complete CAF results from Tab.~\ref{zero-shot} are included as a reference; these use learned fusion coefficients and three-seed averaging, whereas the individual-branch and uniform-fusion controls share one checkpoint.


\begin{table}[!t]
    \centering
    \caption{\textbf{Unlabeled statistics: scope and budget.} Single-seed accuracies (\%) with a fixed prompt checkpoint and fusion coefficients. $N$ denotes the collection size; budgeted references exclude the query image.}
    \label{tab:unlabeled_statistics_budget}
    \small
    \setlength{\tabcolsep}{2.5pt}
    \renewcommand{\arraystretch}{1.02}
    \begin{tabular*}{\columnwidth}{@{\extracolsep{\fill}}llrr@{}}
        \toprule
        Statistics & Available images & Avg-10$\uparrow$ & Avg-4$\uparrow$ \\
        \midrule
        Full collection & All $N$ & 69.524 & 61.649 \\
        Leave-one-out & $N-1$ & 69.530 & \textbf{61.650} \\
        Two-fold cross-fit & Other fold ($\approx N/2$) & \textbf{69.532} & 61.567 \\
        \midrule
        Budget: 256 & $\leq\min(256,N)$ & 68.917 & 60.597 \\
        Budget: 1,024 & $\leq\min(1024,N)$ & 69.210 & 61.056 \\
        Budget: 4,096 & $\leq\min(4096,N)$ & 69.462 & 61.462 \\
        \midrule
        Online cumulative & Past batches & 69.343 & 61.526 \\
        Online EMA & Past batches & 69.287 & 61.543 \\
        Batch leave-one-out & $\leq31$ others & 67.809 & 60.735 \\
        \bottomrule
    \end{tabular*}
\end{table}

\subsection{Scope and Budget of Unlabeled Context}
\label{app:unlabeled_statistics_budget}

Tab.~\ref{tab:unlabeled_statistics_budget} examines how the scope and amount of unlabeled data affect contextual fusion.
We keep the prompt checkpoint and fusion coefficients fixed and vary only the construction of the shared soft-frequency table.
Full-collection statistics use all images in the target dataset.
Leave-one-out statistics exclude the query image, while two-fold cross-fitting estimates each image's context from the opposite fold.
Budgeted variants restrict the number of reference images and also exclude the query.

Excluding the query image leaves performance virtually unchanged: leave-one-out statistics achieve 69.530\% on Avg-10 and 61.650\% on Avg-4, compared with 69.524\% and 61.649\% using the full collection.
Two-fold cross-fitting also retains comparable performance.
Thus, the observed benefit does not depend on including an image in its own context estimate.
Increasing the reference budget from 256 to 4,096 images improves both target averages, with the latter remaining within 0.19 percentage points of full-collection performance on both metrics.

We additionally consider online cumulative and EMA statistics, which use only previously observed batches.
Both predict before incorporating the current batch, using a batch size of 32 and an EMA decay of 0.99 where applicable.
These online variants remain within 0.24 percentage points of full-collection performance on both averages in the recorded evaluation order.
In contrast, estimating context solely from the other images in the current batch reduces Avg-10 to 67.809\% and Avg-4 to 60.735\%.
Together, these results indicate that accumulating context across images is beneficial, while useful contextual estimates can still be obtained with limited reference data or sequential access.

\begin{figure*}[!p]
    \centering
    \includegraphics[width=\textwidth]{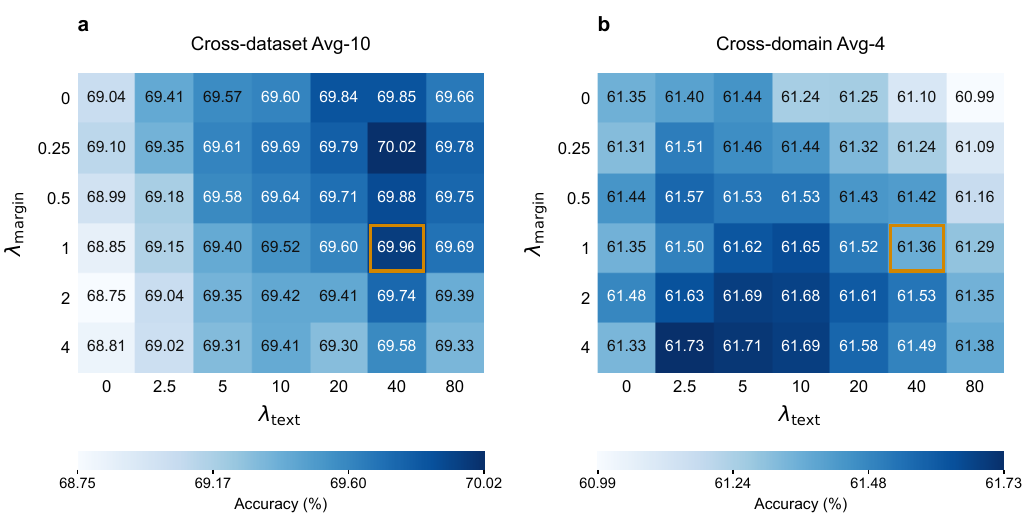}
    \caption{\textbf{Regularization-weight results.} Accuracy (\%) for (a) cross-dataset and (b) cross-domain transfer, with separate color scales. Results combine historical fixed-coefficient and learned-coefficient runs. Outlined cells at $(1,40)$ report the three-seed means from Tab.~\ref{zero-shot}; other cells report individual runs.}
    \label{fig:regularization_sweep}
\end{figure*}

\begin{figure*}[!p]
    \centering
    \includegraphics[width=\textwidth]{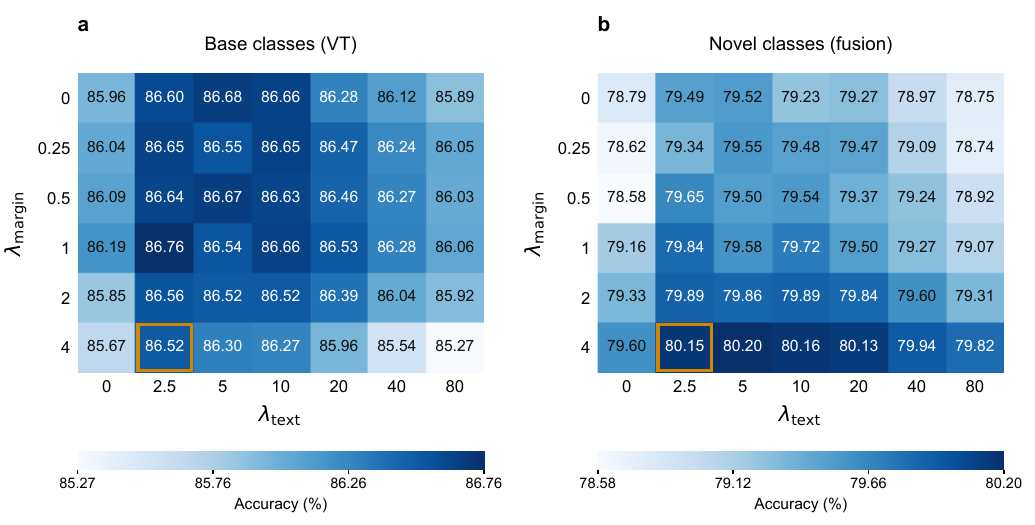}
    \caption{\textbf{Regularization weights for base-to-novel generalization.} Accuracy (\%) averaged over ten datasets and three seeds: (a) base classes using the $VT$ branch and (b) novel classes using fusion, with separate color scales. Outlined cells mark the shared $(4,2.5)$ setting used for CAF in Tab.~\ref{tab:base2novel_style_match}.}
    \label{fig:b2n_regularization_sweep}
\end{figure*}
\clearpage
\endgroup

\subsection{Regularization-Weight Results}
\label{app:regularization_sweep}

Figs.~\ref{fig:regularization_sweep} and~\ref{fig:b2n_regularization_sweep} examine 42 combinations of regularization weights, with $\lambda_{\mathrm{margin}}\in\{0,0.25,0.5,1,2,4\}$ and $\lambda_{\mathrm{text}}\in\{0,2.5,5,10,20,40,80\}$.
The grid in Fig.~\ref{fig:regularization_sweep} combines runs with fixed and learned fusion coefficients. The outlined $(1,40)$ configuration reports the three-seed CAF means from Tab.~\ref{zero-shot}, while the other cells report individual runs.

In Fig.~\ref{fig:regularization_sweep}, the $(1,40)$ configuration adopted in Tab.~\ref{zero-shot} achieves 69.96\% on the ten-dataset cross-dataset average (Avg-10), within 0.07 percentage points of the grid maximum (70.02\%), alongside 61.36\% on the four-variant cross-domain average (Avg-4).
Shifting to $(4,2.5)$ raises Avg-4 to its peak of 61.73\% while reducing Avg-10 to 69.02\%.
This trade-off is consistent with the distinct nature of the two evaluation tracks. When transferring from the 1,000-class ImageNet source to new target vocabularies (Avg-10), strong textual anchoring ($\lambda_{\mathrm{text}}=40$) may help preserve generalizable class semantics. Under visual shifts among classes drawn from the source vocabulary (Avg-4), the highest accuracy occurs with stronger regularization of reliable frozen classification margins ($\lambda_{\mathrm{margin}}=4$).

Fig.~\ref{fig:b2n_regularization_sweep} reveals a complementary trade-off between base-class fitting and novel-class generalization in the few-shot base-to-novel setting, where we evaluate the $VT$ branch on base classes and adaptive fusion on novel classes across ten datasets and three seeds.
Novel-class accuracy consistently peaks at a stronger margin weight of $\lambda_{\mathrm{margin}}=4$ across all tested text weights, whereas base-class accuracy peaks at a milder $(1,2.5)$ setting. This pattern is consistent with margin regularization limiting visual overfitting to the few labeled examples per base class.
Within the $\lambda_{\mathrm{margin}}=4$ regime, increasing $\lambda_{\mathrm{text}}$ from 2.5 to 5 yields a small novel-class gain (80.15\% to 80.20\%) at the cost of a larger drop in base accuracy (86.52\% to 86.30\%).
Consequently, the $(4,2.5)$ setting secures near-optimal novel-class transfer while preserving strong base-class discrimination, achieving the highest harmonic mean of the ten-dataset base and novel averages (83.21\%) across all 42 configurations.

\nextsupplementpart{C}
\section{Analysis Experiments}
\label{app:additional_analyses}


\begin{table}[!t]
    \centering
    \caption{\textbf{Dynamic weight assignment interventions.} Interventions use Branch-soft weights from one checkpoint; shuffled rows average 20 permutations. Shared-soft CAF is a separate reference. Accuracies are percentages.}
    \label{tab:dynamic_weight_interventions}
    \small
    \setlength{\tabcolsep}{3pt}
    \renewcommand{\arraystretch}{1.12}
    \begin{tabular*}{\columnwidth}{@{\extracolsep{\fill}}lrr@{}}
        \toprule
        Weight assignment & Avg-10$\uparrow$ & Avg-4$\uparrow$ \\
        \midrule
        Original (Branch-soft) & 69.405 & 61.641 \\
        Dataset-mean weights & 67.670 & 61.097 \\
        Predicted-class mean weights & 68.058 & 61.137 \\
        Shuffle across all images & 67.058 & 60.827 \\
        Shuffle within disagreement samples & 67.071 & 60.666 \\
        \midrule
        \textbf{Shared-soft CAF (reference)} & \textbf{69.524} & \textbf{61.649} \\
        \bottomrule
    \end{tabular*}
\end{table}

\begin{figure*}[!t]
    \centering
    \begin{minipage}[t]{\columnwidth}
    \vspace{0pt}\centering
    \includegraphics[width=\linewidth]{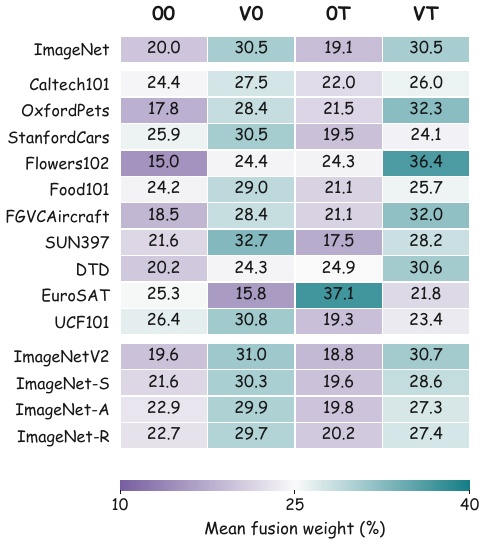}
    \caption{\textbf{Mean CAF fusion weights for cross-dataset and cross-domain evaluation.} Values (\%) are averaged over test images in each dataset, with 25\% corresponding to uniform weighting.}
    \label{fig:mean_fusion_weights}
    \end{minipage}\hfill
    \begin{minipage}[t]{\columnwidth}
    \vspace{0pt}\centering
    \includegraphics[width=\linewidth]{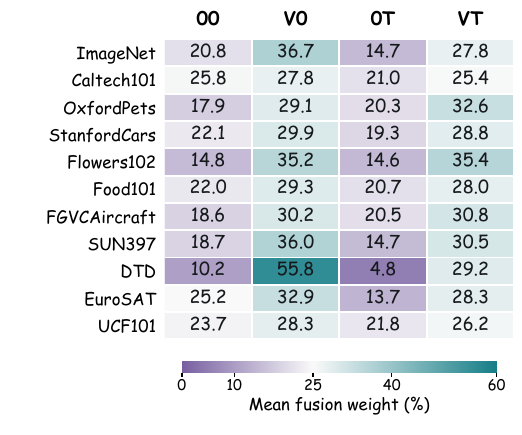}
    \caption{\textbf{Mean fusion weights on novel classes across eleven datasets.} CAF$^{*}$ uses the dataset-specific loss weights from Tab.~\ref{tab:base2novel_style_match}. Values are mean branch weights (\%) on each novel-class test split. Uniform weighting assigns 25\% to each branch.}
    \label{fig:novel_fusion_weights}
    \end{minipage}
\end{figure*}

\subsection{Interventions on Dynamic Weight Assignment}
\label{app:dynamic_weight_interventions}

To examine the importance of matching fusion weights to individual images, Tab.~\ref{tab:dynamic_weight_interventions} modifies weight assignments while keeping the branch logits and frequency statistics fixed.
These interventions use the Branch-soft variant as the unmodified baseline; Shared-soft CAF is included separately for reference.
All controls operate within each dataset without retraining or using target labels.

We consider four interventions.
Dataset-mean weights replace each image's weights with the average vector estimated from the opposite fold of a two-fold split.
Predicted-class mean weights instead average within groups defined by the top-1 prediction of uniform fusion, using the opposite fold and falling back to its global mean when a group is absent.
Shuffle across all images randomly permutes complete four-dimensional weight vectors, preserving their distribution while disrupting their correspondence with images.
Shuffle within disagreement samples applies this permutation only to images whose branches predict different classes.
Both shuffling experiments average results over 20 permutations.

The original assignment outperforms all four controls on both target averages.
Replacing image-specific weights with dataset means reduces Avg-10 and Avg-4 by 1.74 and 0.54 percentage points, respectively.
Conditioning the mean weights on predicted classes partially recovers performance but remains below the original assignment.
Shuffling across all images causes larger reductions of 2.35 and 0.81 percentage points, despite preserving the distribution of weight vectors.
Performance also decreases when shuffling is restricted to disagreement samples.
These results support the value of assigning weights to the images for which they were computed, beyond retaining dataset-level or predicted-class-level weighting preferences.

\subsection{Dataset-Wise Fusion Weights}
\label{app:mean_fusion_weights}

Fig.~\ref{fig:mean_fusion_weights} shows how the same ImageNet-trained CAF model distributes its mean fusion weights across evaluation benchmarks.
On ImageNet and its four cross-domain variants, the visually prompted branches $V0$ and $VT$ consistently receive the two largest shares (29.7\%--31.0\% for $V0$ and 27.3\%--30.7\% for $VT$). Among the variants, the share of the frozen $00$ branch ranges from 19.6\% on ImageNetV2 to 22.9\% on ImageNet-A.
Across the ten cross-dataset targets, three patterns emerge. First, the jointly prompted $VT$ branch leads on Flowers102 (36.4\%), OxfordPets (32.3\%), FGVCAircraft (32.0\%), and the texture benchmark DTD (30.6\%). Second, $V0$, which pairs adapted visual features with frozen text prototypes, receives the largest share on SUN397 (32.7\%), UCF101 (30.8\%), StanfordCars (30.5\%), Food101 (29.0\%), and Caltech101 (27.5\%). On these datasets, the leading branch thus retains the pretrained text representation of each target class. Third, the preference reverses on EuroSAT satellite imagery, where $V0$ receives 15.8\% and $0T$ receives 37.1\%.
Even with fixed model parameters, the balance between visual and textual adaptation therefore varies across target domains.

Fig.~\ref{fig:novel_fusion_weights} reports mean branch weights for CAF$^{*}$ on novel classes after base-class training on each of the eleven datasets.
Here the visual prompts are learned within the target visual domain, while the textual prompts are learned using only the base categories. Across all eleven benchmarks, $V0$ and $VT$ jointly receive more than half of the total weight (53.1\%--85.0\%). The text-only adapted branch $0T$ receives 4.8\%--21.8\% and has the smallest share on nine datasets. On OxfordPets and FGVCAircraft, $00$ has the smallest share.
The concentration of the weights also varies by task. On Caltech101, all four shares are close to the uniform weight of 25\% (21.0\%--27.8\%). On DTD, by contrast, $V0$ alone receives 55.8\%, compared with 4.8\% for $0T$. ImageNet (36.7\% versus 14.7\%) and SUN397 (36.0\% versus 14.7\%) show the same preference for in-domain visual prompts paired with frozen text prototypes.
EuroSAT illustrates how the leading branch can change on the same benchmark across transfer settings. With ImageNet-trained prompts, $0T$ leads at 37.1\% in Fig.~\ref{fig:mean_fusion_weights}. With prompts trained on EuroSAT base classes and evaluated on its novel classes, $V0$ leads at 32.9\%, while $0T$ receives 13.7\% in Fig.~\ref{fig:novel_fusion_weights}.

\end{document}